\documentclass{article}
\usepackage[final]{colm2026_conference}

\usepackage{microtype}
\usepackage{amsmath}
\usepackage{graphicx}
\usepackage{hyperref}
\usepackage{url}
\usepackage{booktabs}
\usepackage{lineno}
\usepackage{xcolor}
\usepackage{xspace}
\usepackage{tabularx}
\usepackage{enumitem}
\usepackage{amssymb}
\usepackage{mdframed}
\usepackage{float}
\usepackage{makecell}

\definecolor{darkblue}{rgb}{0, 0, 0.5}
\hypersetup{colorlinks=true, citecolor=darkblue, linkcolor=darkblue, urlcolor=darkblue}

\usepackage{bm}
\usepackage{xcolor}
\definecolor{darkgreen}{rgb}{0, 0.5, 0}

\newcommand{\xmode}{\textsc{XMode}\xspace} 
\newcommand{\modelresp}{\textsc{ReSP}\xspace} 
\newcommand{\ours}{\textsc{LakeQuest}\xspace} 
\newcommand{\numexamples}{9,846\xspace} 
\newcommand{\numexamplesshort}{9.8k\xspace} 

\newcolumntype{Y}{>{\centering\arraybackslash}X}

\title{\ours{}: A Three-Domain Benchmark for Grounded \\Question Answering across Data Lakes}

\newcommand{\waterloo}{\raisebox{3pt}{\includegraphics[height=0.8em]{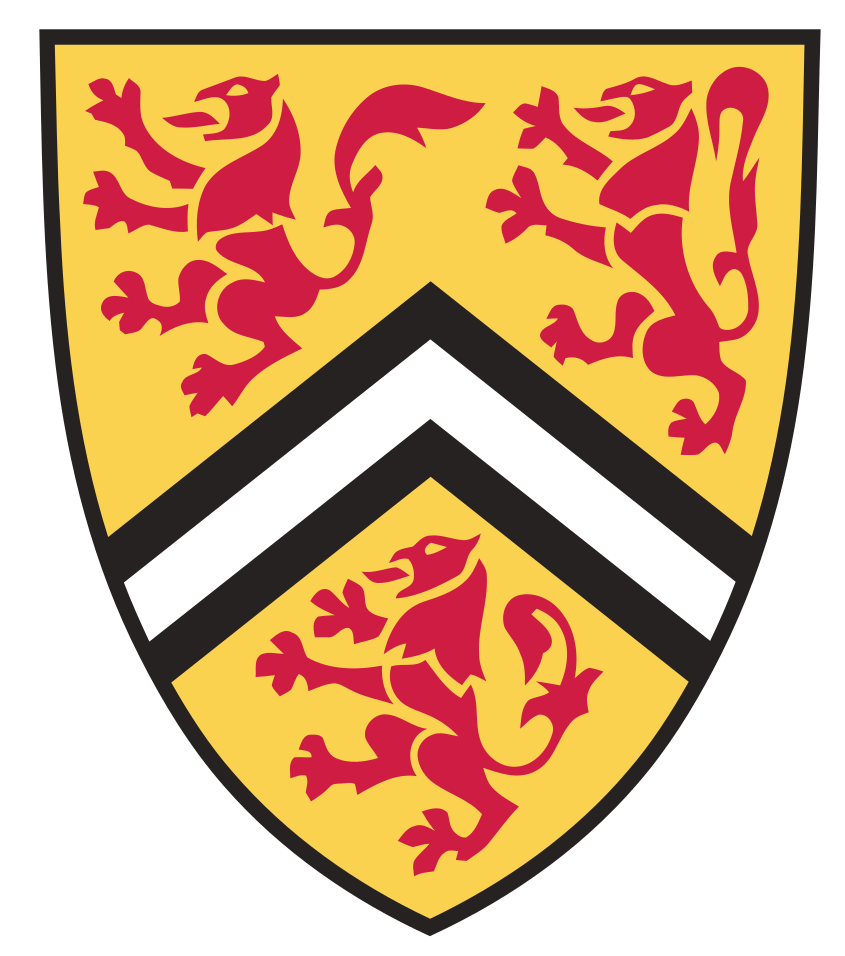}}}
\newcommand{\lsix}{\raisebox{3pt}{\includegraphics[height=0.7em]{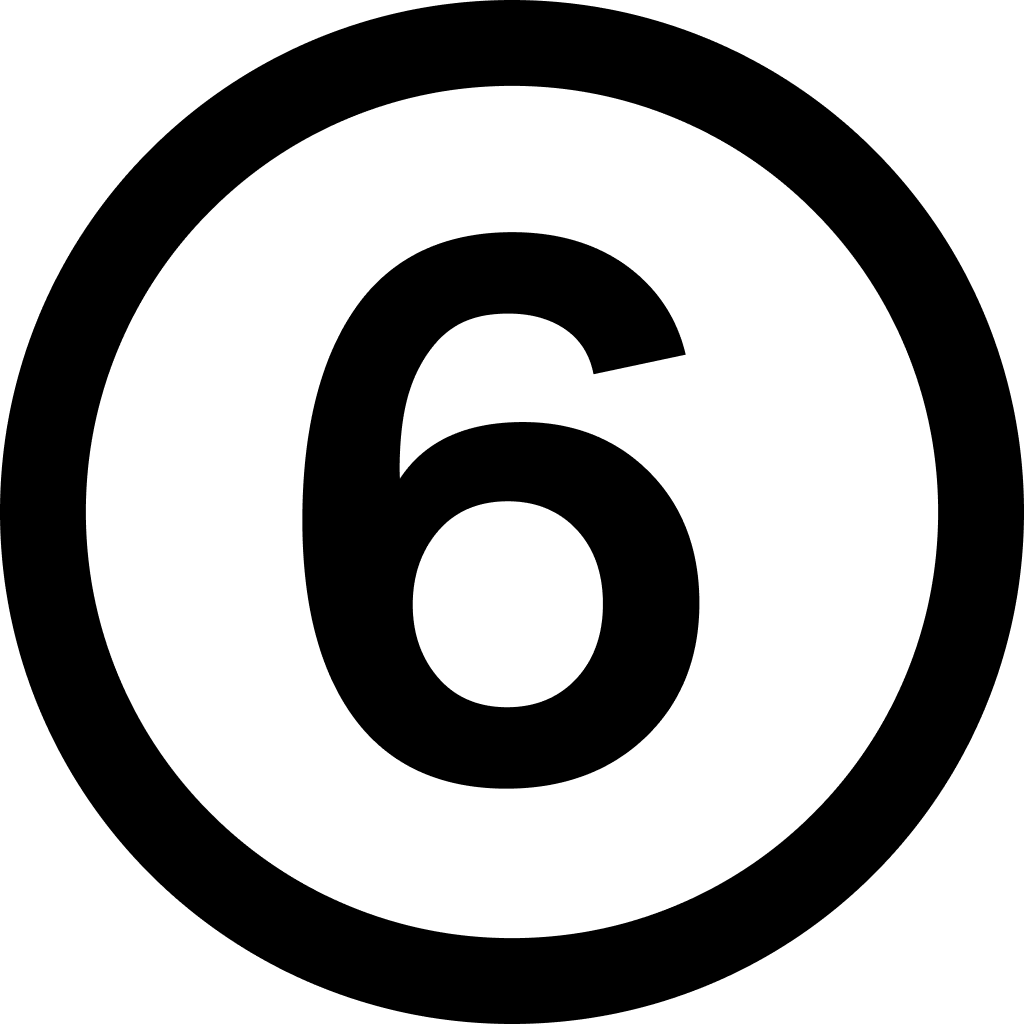}}}

\newcommand{\githubdown}{\includegraphics[height=0.8em]{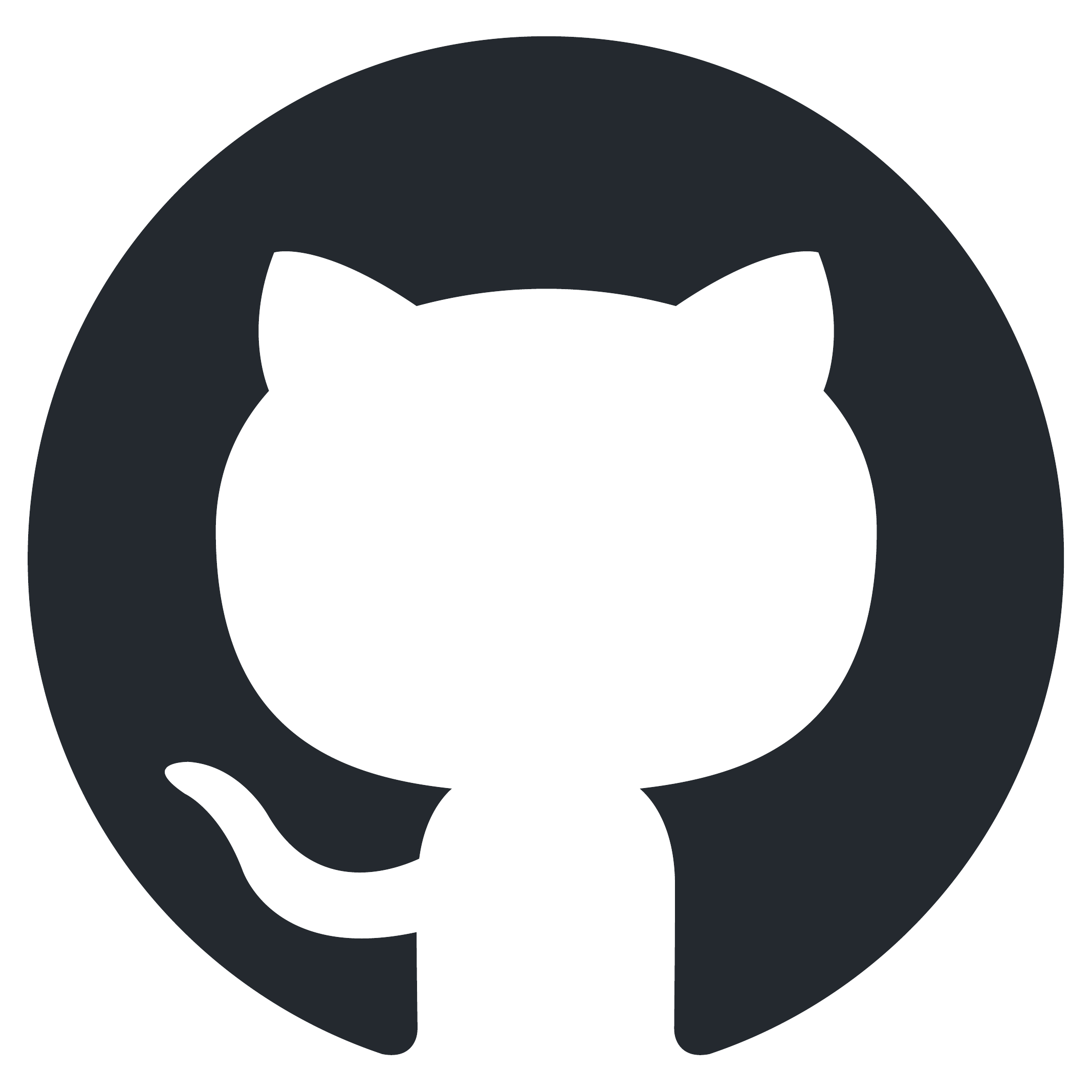}\xspace}

\author{Michael Solodko\waterloo\thanks{{Correspondence to: \texttt{msolodko@uwaterloo.ca}}}\quad Steven Gong\waterloo\quad Guangwei Yu\lsix\quad Satya Krishna Gorti\lsix\\ \textbf{Jesse C. Cresswell}\lsix\quad \textbf{Victor Zhong}\waterloo
 \\
\centerline{\waterloo University of Waterloo \quad \lsix Layer 6 AI}
\\\\
\githubdown Code, data, and starter kit:
\url{https://github.com/michael0402/LakeQuest}
\\
LakeQuest Homepage:
\url{https://michael0402.github.io/LakeQuest/}
}

\begin{document}
\ifcolmsubmission
\linenumbers
\fi

\maketitle

\begin{abstract}
While modern question answering (QA) systems excel on clean, schema-aligned corpora, real-world knowledge is rarely so neatly packaged.
Answering questions over enterprise and scientific data lakes requires systems to navigate heterogeneous, weakly structured collections of tables, passages, and linked metadata.
Current benchmarks abstract away this noisy discovery process, failing to evaluate end-to-end performance.
To bridge this gap, we introduce \ours, a human-validated benchmark of \numexamples{} QA pairs designed to evaluate the end-to-end retrieve-and-synthesize pipeline over realistic data lakes.
\ours{} spans three diverse domains---AI/ML metadata, retail banking, and multimodal biomedical drug information---and pairs every question with exact, modality-aware evidence pointers. By isolating source discovery from cross-modal synthesis, \ours{} exposes critical failure modes in modern QA systems.
Our baseline evaluations, including standard Retrieval-Augmented Generation (RAG) and agentic tool-use methods, reveal that high-quality retrieval does not guarantee correct reasoning.
Systems consistently struggle with relation chaining in metadata graphs, policy grounding in bank ledgers, and joint tabular QA in biomedical contexts, highlighting the need for robust discovery and faithful cross-file composition mechanisms in future agentic QA systems.
\end{abstract}

\section{Introduction}
\vspace{-4pt}
Question answering (QA) systems have achieved strong performance on curated, homogeneous corpora, ranging from reading comprehension \citep{rajpurkar2016squad} to open-domain text retrieval \citep{kwiatkowski2019naturalquestions} and multi-hop reasoning \citep{yang2018hotpotqa}.
However, real-world enterprise and scientific knowledge is rarely stored as flat text. Instead, it often resides in \textbf{data lakes}: heterogeneous, weakly structured collections of tables, passages, and linked metadata.

Answering natural language questions over data lakes requires navigating siloed sources, combining cross-modal evidence, and maintaining explicit provenance.
However, modern retrieve-and-synthesize architectures \citep{lewis2020rag, yao2023react} lack evaluation environments that simultaneously stress discovery, cross-source composition, and faithful attribution over heterogeneous corpora.
Existing benchmarks bypass the noise, scale, and weak metadata characteristic of real data lakes, evaluating either pure data discovery without natural language synthesis, or reasoning over already-clean, schema-aligned text.

To bridge this gap, we introduce \ours, a benchmark designed to evaluate the end-to-end QA pipeline over realistic data lakes.
\ours provides a shared construction and evaluation framework instantiated across three diverse domains: AI/ML metadata, a realistic retail bank, and multimodal biomedical drug information.
Rather than flattening these domains into a uniform representation, \ours preserves their native structural heterogeneity as data lakes.
Each benchmark instance pairs a natural language question with a reference answer and exact, modality-aware evidence pointers (e.g., specific table rows or passage spans).
Furthermore, because \ours preserves the native structure of the data lakes, it inherently captures factual redundancy. A system might successfully synthesize a grounded answer by discovering valid \emph{alternative provenance} (e.g., a redundant table or densely informative passage) even if the target gold artifact is missed.
This explicit attribution enables precise isolation of system failures into discovery errors (failing to retrieve evidence) versus reasoning errors (failing to synthesize an answer from the retrieved evidence) \citep{bryan2025taxonomy, leung2026classifying}.

We construct \ours{} using a reproducible five-stage pipeline encompassing domain instantiation, language-model-driven QA synthesis, structural filtering, semantic validation, and rigorous human review. The resulting benchmark contains \numexamples human-validated QA pairs that entail single-source lookup, cross-source composition, and multi-hop reasoning.

Evaluating a diverse suite of baselines---from zero-shot reasoning \citep{kojima2022zeroshotcot} and standard retrieval-augmented generation (RAG) \citep{lewis2020rag} to structured pipelines and agentic tool-use methods such as \xmode{} \citep{nooralahzadeh2025m2ex} and \modelresp{} \citep{jiang2025resp}---we find that the bottlenecks in data lake QA remain fundamentally unsolved and are strongly domain-dependent.
Specifically, accurate retrieval does not guarantee faithful reasoning when cross-modal composition is required, and even agentic strategies struggle with multi-hop discovery over weakly structured metadata.

Our primary contributions are: \textbf{(1) The \ours{} Benchmark:} A human-validated dataset of over \numexamplesshort{} questions evaluating grounded reasoning across three heterogeneous data lakes (AI/ML, Retail Banking, and Biomedical Drug). \textbf{(2) A Reproducible Construction Framework:} A five-stage protocol for synthesizing, filtering, and validating QA pairs with explicit evidence traces over weakly structured corpora. \textbf{(3) Empirical Analysis of Modern QA:} Extensive baseline evaluations revealing that high retrieval quality does not consistently translate into accurate, faithful answers---especially when multi-step or cross-source composition is required.

\begin{figure}[t]
\centering
\includegraphics[width=\linewidth]{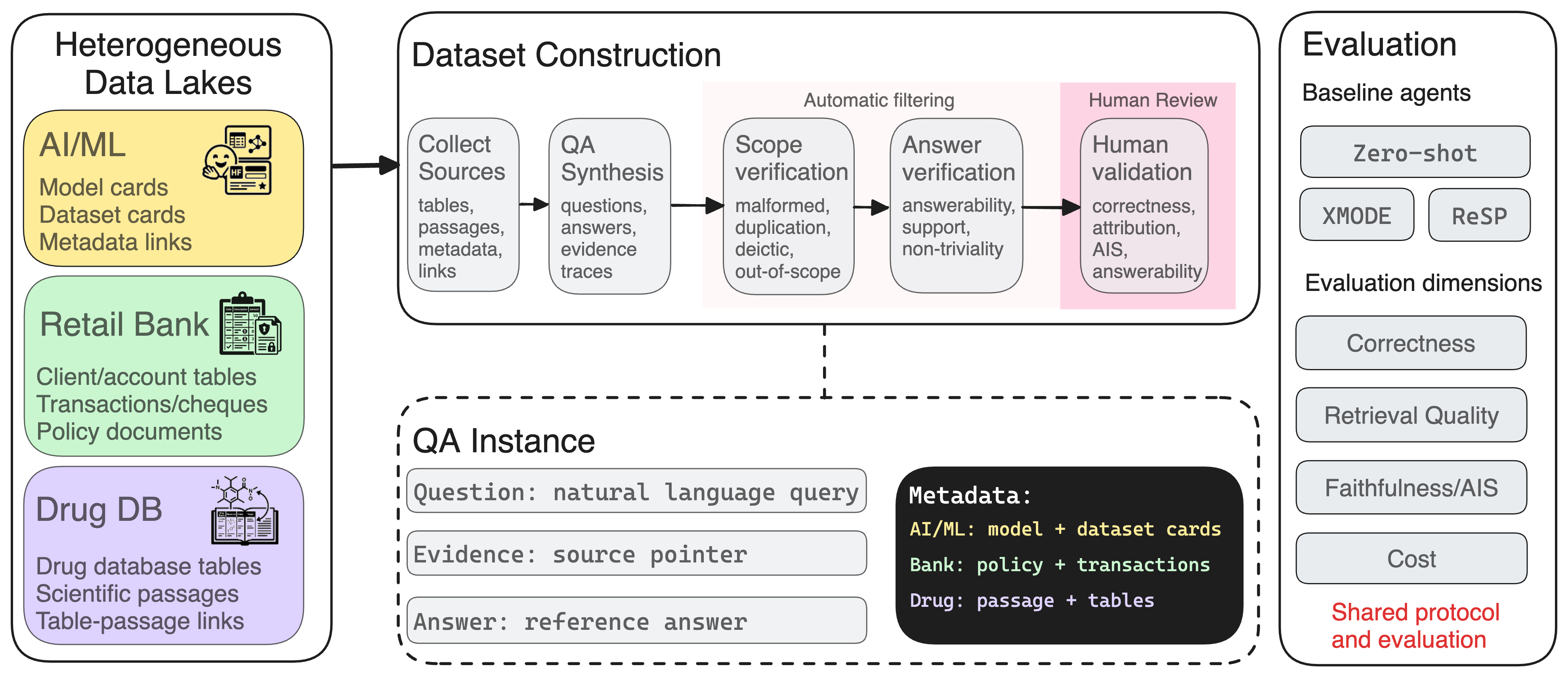}
\vspace{-0.2in}
\caption{\small Overview of \ours{}. The benchmark spans three heterogeneous data lakes built under a shared construction and evaluation protocol for grounded question answering.}
\label{fig:overview}
\vspace{-0.2in}
\end{figure}

\section{Related Work}
\label{sec:related}
\vspace{-4pt}

Table~\ref{tab:comparison} provides a structural comparison between \ours{} and existing QA and data discovery benchmarks across modalities, evidence tracing, and evaluation scope.

\begin{table}[th]
\centering
\small
\setlength{\tabcolsep}{3.0pt}
\renewcommand{\arraystretch}{1.0}
\renewcommand{\tabularxcolumn}[1]{m{#1}}
\caption{\small Comparison of \ours{} with existing QA and data discovery benchmarks. \textbf{Modality} describes the data formats involved. \textbf{Ev} indicates the presence of explicit evidence traces linking answers to sources. \textbf{Md} denotes the inclusion of weak metadata and distractor documents to test retrieval robustness. \textbf{Val} indicates datasets that underwent human validation. \textbf{Objects} denotes approximate source-side evidence containers (e.g., paragraphs, tables, passages, or hybrid contexts).}\label{tab:comparison}
\begin{tabularx}{\linewidth}{>{\raggedright\arraybackslash}X Y c c c c c Y Y}
\toprule
\textbf{Benchmark} & \textbf{Modality} & \textbf{Ev} & \textbf{Md} & \textbf{Val} & \textbf{\# Qs} & \textbf{Objects} & \textbf{Task} & \textbf{Domain} \\
\midrule
HotpotQA & Text & \textcolor{darkgreen}{$\bm{\checkmark}$} & \textcolor{red}{$\bm{\times}$} & \textcolor{darkgreen}{$\bm{\checkmark}$} & 113k & \makecell[c]{5M+\\paragraphs} & \makecell[c]{Text\\Reasoning} & Wikipedia \\
\midrule
HybridQA & \makecell[c]{Tables\\Text} & \textcolor{red}{$\bm{\times}$} & \textcolor{red}{$\bm{\times}$} & \textcolor{darkgreen}{$\bm{\checkmark}$} & 69k+ & \makecell[c]{13k tables +\\293k passages} & \makecell[c]{Hybrid\\Reasoning} & Wikipedia \\
\midrule
TAT-QA & \makecell[c]{Tables\\Text} & \textcolor{red}{$\bm{\times}$} & \textcolor{red}{$\bm{\times}$} & \textcolor{darkgreen}{$\bm{\checkmark}$} & 16k+ & \makecell[c]{2.5k+ hybrid\\contexts} & \makecell[c]{Numerical\\QA} & Finance \\
\midrule
LakeBench & \makecell[c]{Tables\\Metadata} & \textcolor{red}{$\bm{\times}$} & \textcolor{darkgreen}{$\bm{\checkmark}$} & \textcolor{darkgreen}{$\bm{\checkmark}$} & 10k+ & \makecell[c]{16M+\\tables} & \makecell[c]{Table\\Discovery} & Open Data \\
\midrule
\textbf{\ours} & \makecell[c]{Tables, Text\\Metadata} & \textcolor{darkgreen}{$\bm{\checkmark}$} & \textcolor{darkgreen}{$\bm{\checkmark}$} & \textcolor{darkgreen}{$\bm{\checkmark}$} & 9.8k &
\makecell[c]{1M+ model/data cards + \\ 10k+ bank records + \\ 500k+ drug texts/tables}
& \shortstack{Grounded\\QA} & \shortstack{AI/ML \\ Banking \\ Drug} \\
\bottomrule
\end{tabularx}
\vspace{-8pt}
\end{table}

\vspace{-4pt}
\textbf{Data Discovery and Integration.} Data management literature extensively studies how to discover, link, and integrate datasets \citep{fernandez2018aurum, fan2023starmie}. Accordingly, benchmarking efforts in this space, such as LakeBench \citep{deng2024lakebench} and CMDL \citep{eltabakh2023cmdl}, evaluate the retrieval of joinable tables or cross-modal artifacts. While essential for data discovery, these benchmarks remain purely retrieval-focused and do not evaluate natural language synthesis.

\textbf{Text-centric and Multi-hop QA.}
In the NLP community, QA benchmarks have evolved from single-document reading comprehension \citep{rajpurkar2016squad} to open-domain retrieval \citep{kwiatkowski2019naturalquestions} and multi-hop reasoning with HotpotQA \citep{yang2018hotpotqa}.
While foundational, these datasets largely assume text-centric evidence within curated, relatively homogeneous corpora (e.g., Wikipedia).
Real data lakes, however, are characterized by weak metadata, fragmented entity references, and a mixture of semi-structured objects, making reasoning pathways noisier than those found in standard text-only multi-hop settings.

\textbf{Hybrid and Multimodal QA.} A closer line of work evaluates QA over multiple evidence types.
HybridQA \citep{chen2020hybridqa} and OTT-QA \citep{chen2021ottqa} require reasoning across tables and text, while TAT-QA \citep{zhu2021tatqa} emphasizes numerical reasoning over financial reports. MultiModalQA \citep{talmor2021multimodalqa} expands the scope to joint reasoning over text, tables, and images.
While these benchmarks require cross-modal synthesis, they generally operate over highly curated, schema-aligned, or task-specific corpora.
They do not model the core challenges of enterprise data lakes: navigating siloed, weakly linked sources, resolving entities across heterogeneous files, and maintaining strict, modality-aware provenance for auditability.

\textbf{Retrieval-Grounded and Agentic Systems.} Methodologically, \ours provides a necessary testbed for modern retrieve-and-synthesize pipelines. RAG \citep{lewis2020rag} and fusion-in-decoder (FiD) \citep{izacard2021fid} established paradigms for grounding generation in retrieved contexts, while recent methods propose structured decomposition \citep{zhou2022leasttomost}, iterative tool use \citep{yao2023react, schick2023toolformer}, and multi-agent orchestration like Symphony \citep{tang2024symphony}.
By providing a heterogeneous environment with explicit evidence requirements and realistic discovery bottlenecks, \ours enables the community to rigorously evaluate whether these control strategies can handle the noise and scale of real-world data lakes.

\section{The \ours{} Framework}
\label{sec:framework}
\vspace{-4pt}
To rigorously evaluate QA over heterogeneous data lakes, we require a framework that isolates source discovery from cross-modal synthesis, scales across domains, and maintains strict provenance.
\ours{} operationalizes this through a formalized task definition and a reproducible construction pipeline.

\subsection{Problem Formulation and Task Taxonomy}
\vspace{-4pt}
We formulate grounded data lake QA: given a data lake $\mathcal{D} = \{d_1, \dots, d_n\}$ (where $d_i$ is a table, passage, or metadata record) and a natural language question $q$, a system $f$ must return a textual answer $\hat{a}$ and a precise supporting evidence set $\hat{E} \subseteq \mathcal{D}$: $f(q, \mathcal{D}) \rightarrow (\hat{a}, \hat{E})$.
Modern agents decompose this operation into retrieve-and-synthesize workflows: $\hat{E} = R(q, \mathcal{D})$ and $\hat{a} = G(q, \hat{E})$.
By requiring systems to output $\hat{E}$, \ours{} allows us to independently evaluate the retrieval mechanism $R$ (discovery) and the generator $G$ (synthesis).

To systematically test this pipeline, \ours{} is organized into four distinct reasoning families: \textbf{single-source lookups} requiring fact extraction from isolated objects; \textbf{cross-source composition} necessitating the combination of facts across multiple heterogeneous objects (e.g., joining a policy with a transaction); \textbf{multi-hop reasoning} requiring chained inference over intermediate entities or linked metadata; and \textbf{anchor-based retrieval}, which conditions questions on a selected entity, forcing the agent to reason over related entities amidst semantically close distractors.

Unlike standard text QA benchmarks where evidence is simply a document ID, \ours{} enforces modality-aware provenance. Each benchmark instance is defined by the tuple $x_i = (q_i, a^*_i, E^*_i, m_i)$, where $E^*_i$ resolves to concrete sub-document units (e.g., specific table rows or exact passage spans) and $m_i$ tracks the question family and domain metadata. Figure~\ref{fig:instance_anatomy} illustrates a concrete cross-source instance from the Retail Bank data lake.

\begin{figure}[t]
\centering
\fbox{
\begin{minipage}{0.95\textwidth}
\small
\textbf{Domain:} Retail banking \hfill \textbf{Family:} Cross-source composition (Policy + Table + Metadata) \\
\rule{\linewidth}{0.4pt}
\textbf{Question ($q_i$):} Based on standard routing policies, does the wire transfer initiated by Acme Corp on Oct 14th require manual manager approval? \\
\textbf{Gold Evidence ($E^*_i$):}
\begin{itemize}[leftmargin=*, nosep]
    \item \texttt{Policy\_Manual\_v3}: Span [1124-1180] \textit{(``...Tier 2 business accounts require manual manager approval for all outbound international wire transfers exceeding \$15,000...'')}
    \item \texttt{Transaction\_Ledger\_72} (Acme Corp): Row 84, Columns \texttt{Type}, \texttt{Amount} \textit{(``Int\_Wire  \$18,500.00'')}
    \item \texttt{Client\_Profile\_7742}: Key \texttt{Account\_Tier} \textit{(``Tier 2'')}
\end{itemize}
\rule{\linewidth}{0.4pt}
\textbf{Reference Answer ($a^*_i$):} Yes, the \$18,500 transfer exceeds the \$15,000 limit for Tier 2 accounts.
\end{minipage}
}
\caption{\small Anatomy of a \ours{} instance --- this requires identifying and synthesizing specific, sub-document evidence units across multiple heterogeneous sources to derive the correct answer.}
\label{fig:instance_anatomy}
\vspace{-8pt}
\end{figure}

\subsection{The Shared Five-Stage Pipeline}
\vspace{-4pt}
Manually authoring tens of thousands of complex, cross-modal QA pairs is unscalable, while purely synthetic datasets often suffer from hallucinations and trivial reasoning pathways.
To balance scale with rigor, we construct \ours{} using an LLM-in-the-loop pipeline consisting of five shared stages across all domains (a visual schematic of the Retail Bank data lake question-answer synthesis process is provided in App~\ref{app:finlake_qa_generation}).

The five shared stages include: \textbf{(1) Lake Instantiation}, preserving the domain's native schema rather than homogenizing documents; \textbf{(2) Candidate Synthesis}, where a strong language model (OpenAI GPT-5, Feb. 1, 2026) synthesizes a candidate question $q$, answer $a^*$, and exact evidence pointers $E^*$ from a sampled evidence subgraph; \textbf{(3) Structural Filtering}, which programmatically removes malformed outputs or deictic questions; \textbf{(4) Semantic Filtering}, utilizing an LLM-as-a-judge to reject trivial candidates answerable via parametric memory alone; and \textbf{(5) Human Validation}, the definitive gate for benchmark inclusion, assessing answerability and attribution. Exact prompts and rubrics are detailed in App~\ref{app:prompts}.

\subsection{Human Validation and Quality Control}
\label{sec:human-eval}
\vspace{-4pt}
To guarantee evaluation fidelity, all items passing the automated filters are subjected to rigorous human validation. Annotators were tasked with independently verifying two criteria: \emph{Answerability} (can the question be answered exclusively using the cited evidence?) and \emph{Attribution} (is the generated answer completely faithful to the cited text/table?).
The use of human annotators was approved by the ethics review board at the authors' institution. Prolific annotators (paid \$12 USD/hr) independently verified each item under this protocol. Inter-annotator agreement (Fleiss' $\kappa$) demonstrates strong consistency across raters: $0.82$ for Answerability and $0.85$ for Attribution.
Full annotator guidelines, including a screenshot of the custom validation interface (Figure~\ref{fig:annotation_ui}), are provided in App~\ref{app:annotation}.
Only instances that achieved strict human consensus were retained in the final \ours{} release.

\begin{table}[t]
\centering
\small
\caption{\small Abstraction map to unify task design while preserving domain-native structure.}
\label{tab:domain_abstractions}
\begin{tabularx}{\linewidth}{@{} >{\raggedright\arraybackslash}p{0.06\linewidth} >{\raggedright\arraybackslash}X >{\raggedright\arraybackslash}X >{\raggedright\arraybackslash}X @{}}
\toprule
\textbf{Domain} & \textbf{Lake Representation} & \textbf{Task Abstraction} & \textbf{Primary Reasoning Stress} \\
\midrule
AI/ML & Linked metadata entities (models, datasets, relations) & Entity/relation QA with lookup, composition, hop, and anchor families & Entity disambiguation and relation chaining \\
\midrule
Retail Bank & Structured operational records with policy context & Persona-workflow tasks with client/group scopes & Decision grounding under workflow/policy constraints \\
\midrule
Drug & Hybrid tables + scientific passages with linkage artifacts & Evidence-bundle QA across single-modality and mixed-modality settings & Cross-modality evidence integration and multi-hop synthesis \\
\bottomrule
\end{tabularx}
\vspace{-8pt}
\end{table}

\section{The \ours{} Dataset --- Instantiating the Three Data Lakes}
\label{sec:lakes}
\vspace{-4pt}
To evaluate the framework across diverse reasoning requirements, \ours{} instantiates three data lakes: an AI/ML metadata repository, a synthetic retail banking system, and a biomedical corpus.
Table~\ref{tab:domain_abstractions} outlines the high-level abstractions used to preserve the native structures of these domains.
Furthermore, Table~\ref{tab:domain_status} details the final benchmark coverage and human annotation outcomes across these three lakes.

\paragraph{AI/ML Data Lake}
The AI/ML data lake focuses on entity disambiguation and relation chaining over weakly linked metadata.
We construct the corpus using a snapshot of the Hugging Face Hub API \citep{hfhubapi}.
The resulting lake consists of model cards and dataset cards, which are inherently semi-structured.
We instantiate 5 distinct question families in this domain, yielding 3,578 human-validated questions (see Table~\ref{tab:domain_status} for a high-level summary and Table~\ref{tab:aiml_human_results} in App~\ref{app:statistics} for the detailed breakdown).

These tasks map to our core reasoning taxonomy by requiring systems to deduce specific hyperparameters from isolated cards (\textbf{single-source}), synthesize constraints by jointly reasoning over explicitly linked model and dataset cards (\textbf{cross-source}), chain relations across the metadata graph to link datasets and subsequent models (\textbf{multi-hop}), and formulate queries around specific anchor entities despite lexical drift (\textbf{anchor-based}).

\paragraph{Retail Bank Data Lake}
The Retail Bank data lake evaluates grounded decision-making under strict policy constraints. 
Because real operational banking data is highly restricted, we generate a high-fidelity synthetic lake by deriving generation seed examples and instantiating an open-banking-inspired relational schema (client, account, transaction, and policy entities). We then populate client profiles, transaction ledgers, and institutional policy documents.
We further validate that transaction distributions are similar between the synthetic lake and original real banking data by matching graph density on client profiles.
Tasks are formulated across 9 persona-driven workflows (e.g., a fraud analyst investigating a transaction, or an auditor reviewing compliance), yielding 2,964 validated questions (detailed breakdown provided in Table~\ref{tab:financial_human_results}, App~\ref{app:statistics}).

These workflows map to core reasoning families: \textbf{single-source lookups} for retrieving isolated facts (e.g., account tiers or policy limits); \textbf{cross-source composition} to combine unstructured policy rules with structured transaction ledgers (e.g., checking if a wire transfer violates a tiered policy); \textbf{multi-hop reasoning} to trace complex fund flows across multiple tables before applying operational rules; and \textbf{anchor-based retrieval} to disambiguate compliance policies based on client attributes amidst overlapping regional guidelines.

\paragraph{Drug Data Lake}
The Drug data lake stresses multimodal evidence integration.
We construct this corpus by fusing structured XML records from DrugBank 6.0~\citep{knox2024drugbank6} with open-access scientific literature indexed in PubMed~\citep{sayers2026ncbi}.
This heterogeneous mixture of dense text and structured properties instantiates 5 question families, producing 3,304 validated questions (see Table~\ref{tab:drug_human_results} in App~\ref{app:statistics} for statistical breakdown).

The reasoning families in this domain stress multimodal evidence integration: \textbf{single-source lookups} extract precise values from isolated DrugBank tables or PubMed abstracts; \textbf{cross-source composition} synthesizes answers by linking unstructured text passages (e.g., mechanism of action) to structured tables (e.g., biochemical properties); \textbf{multi-hop reasoning} chains relationships across modalities (e.g., identifying a target in a table, then finding passages detailing its metabolic pathway); and \textbf{anchor-based retrieval} requires identifying correct active compounds amidst structurally similar distractors and complex aliases.

\paragraph{Splits and Licensing}
To prevent contamination, we define (open) validation (20\%) and test (80\%) splits at the \emph{entity} rather than question level (i.e., $\mathrm{Entity}(\mathrm{val}) \cap \mathrm{Entity}(\mathrm{test}) = \emptyset$).
The \ours{} benchmark and underlying corpora are hosted on Hugging Face at \url{anonymousurl}.
The AI/ML and Retail Bank subsets are released under an MIT license. For the Drug subset, we release benchmark annotations and evidence pointers, while use of underlying DrugBank tables remains subject to DrugBank licensing terms and PubMed-linked content remains subject to NLM/PubMed access policies and publisher rights.

\begin{table}[t]
\centering
\small
\caption{\small Benchmark coverage summary across all three lakes after human validation.}
\label{tab:domain_status}
\begin{tabularx}{\linewidth}{@{} l >{\raggedright\arraybackslash}X rrrrr @{}}
\toprule
\textbf{Domain} & \textbf{Question Families} & \textbf{Questions} & \textbf{Correct} & \textbf{AIS} & \textbf{Answerable} \\
\midrule
AI/ML & 5 source combos & 3,935 & 3,750 & 3,687 & 3,578 \\
Retail Bank & 9 personas & 3,312 & 3,047 & 3,003 & 2,964 \\
Drug & 5 source combos & 3,574 & 3,434 & 3,373 & 3,304 \\
\textbf{Total} & \textbf{--} & \textbf{10,821} & \textbf{10,231} & \textbf{10,063} & \textbf{9,846} \\
\bottomrule
\end{tabularx}
\vspace{-8pt}
\end{table}

\section{Experimental Setup and Baselines}
\label{sec:experiments}
\vspace{-4pt}
To benchmark grounded reasoning over data lakes, we define a unified evaluation contract and test representative QA baselines ranging from direct prompting to agentic tool use.

\paragraph{Evaluation Metrics}
\label{subsec:metrics}
Because data lake QA serves both high-assurance settings (where hallucination is unacceptable) and exploratory analytics (where cost and speed are critical), we treat evaluation as a multi-dimensional quality-cost trade-off. 
We evaluate systems across four axes:
\begin{itemize}[leftmargin=*, nosep]
    \item \textbf{Accuracy (Correctness):} We report binary correctness using rubric-based LLM-as-a-judge for baseline runs (including direct-answer and reasoning-answer correctness where applicable), and human-judged correctness for annotated benchmark slices in App~\ref{app:statistics}.
    \item \textbf{Faithfulness (Attribution):} Whether the predicted answer $\hat{a}$ is supported by the predicted evidence set $\hat{E}$. We operationalize this using Attributable to Identified Sources (AIS) style judgments \citep{rashkin2023ais}, aligned with the human attribution protocol used throughout this benchmark.
    \item \textbf{Retrieval Quality:} Domain/modality-aware recall of gold evidence $E^*$ in $\hat{E}$ (e.g., card retrieval in AI/ML, policy and table recall in Retail Bank, and passage/table recall in Drug). For unstructured evidence, success is chunk-level span inclusion; for structured evidence, we require retrieval of the referenced source object and apply row/field-level checks when explicit row-level pointers are available. Because real-world data lakes inherently contain redundant information (e.g., same fact appearing in multiple tables/passages), this metric serves as a strict lower bound. A system may retrieve valid alternative evidence that correctly answers the question despite missing the specific gold artifact.
    \item \textbf{Token Usage:} End-to-end retrieval-plus-synthesis token footprint, reported as total model tokens (and normalized per-query token count) under the evaluation-time configuration.
\end{itemize}

\begin{table}[t]
\centering
\small
\caption{\small Baseline results for the AI/ML data lake. Metrics are Retrieval Recall ($R$), Accuracy ($A$), Faithfulness ($F$), and Token Count ($T$). Token Count ($T$) is reported in thousands of generation tokens per query. Model and Dataset are single-source lookup and anchor questions, Model-Dataset is cross-source composition, and Model-hop and Dataset-hop are multi-hop reasoning questions.}
\label{tab:rag_results_aiml}
\begin{tabularx}{\linewidth}{@{} l *{6}{>{\centering\arraybackslash}X} @{}}
\toprule
& \multicolumn{4}{c}{\textbf{RAG}} & \multicolumn{2}{c}{\textbf{Zero-shot}} \\
\cmidrule(lr){2-5} \cmidrule(l){6-7}
\textbf{Question Type} & $R$ & $A$ & $F$ & $T$ & $A$ & $T$ \\
\midrule
Model         & 0.09 & 0.35 & 0.25 & 0.39 & 0.25 & 0.06 \\
Dataset       & 0.15 & 0.12 & 0.14 & 0.34 & 0.09 & 0.05 \\
Model-Dataset & 0.12 & 0.38 & 0.32 & 0.38 & 0.17 & 0.05 \\
Model-hop     & 0.13 & 0.23 & 0.16 & 0.47 & 0.20 & 0.07 \\
Dataset-hop   & 0.23 & 0.35 & 0.38 & 0.49 & 0.24 & 0.07 \\
\bottomrule
\end{tabularx}
\vspace{-8pt}
\end{table}

\paragraph{Evaluated Baselines}
\label{subsec:baselines}
We establish baseline performance using models with different levels of agentic complexity.
We use GPT-5 as the generator (kept fixed within each experiment suite) \citep{gpt5systemcard2025} and FAISS-backed \texttt{text-embedding-3-small} for retrieval \citep{douze2026faiss}.
Comprehensive details, including system instructions, retrieval formatting templates, and agentic tool-use prompts for the evaluated systems, are provided in App~\ref{app:prompts}.

\begin{itemize}[leftmargin=*, nosep]
    \item \textbf{Zero-Shot Prompting:} A naive baseline that directly prompts the generator LLM without explicit retrieval, testing the limits of parametric memory and instruction following.
    \item \textbf{Retrieval-Augmented Generation (RAG) \citep{lewis2020rag}:} Used as our baseline for the document-only AI/ML data lake, this is a standard retrieve-then-generate pipeline. We chunk unstructured passages and structured metadata cards into overlapping text segments before indexing; we then retrieve the top-$k$ most relevant chunks and synthesize a final answer. We experiment with different values of $k$ and finalize on $k=5$, which consistently attains strong performance.
    \item \textbf{Agentic and Structured Pipelines (\xmode{}, \modelresp{}) \citep{nooralahzadeh2025m2ex,jiang2025resp}:}
    Standard RAG is structurally insufficient for Retail Bank and Drug because it cannot natively execute the relational row operations that navigate Retail Bank transaction records and Drug ledgers. Instead, we evaluate stronger multi-step baselines that go beyond single-shot text retrieval by interleaving reasoning and tool use. In our implementation, \xmode{} acts as a tool-augmented RAG system: it uses an iterative loop with passage retrieval from a FAISS index alongside direct query/SQL access to structured tables. \modelresp{} follows an iterative retrieve--summarize--plan style decomposition.
\end{itemize}

\section{Results and Cross-Domain Analysis}
\label{sec:results}
\vspace{-4pt}
We evaluate baselines across the three domains to understand where modern retrieve-and-synthesize systems fail.
Across domains, retrieval quality alone is insufficient; grounded data-lake QA requires stronger cross-source composition and attribution mechanisms.

\subsection{AI/ML Data Lake: Discovery is the primary bottleneck}
\vspace{-4pt}
The AI/ML lake is challenging due to scale---it is built from a Hugging Face model-card corpus with over 1M cards and normalized into a large model--dataset graph with on-demand card-text retrieval---and the inconsistencies between how different authors document artifacts.
As shown in Table~\ref{tab:rag_results_aiml}, the RAG baseline exhibits a notable gap between exact-match Recall ($R$) and Accuracy ($A$). For example, on isolated Model questions, RAG achieves 0.35 Accuracy despite a recall of 0.09. This gap validates that real data lakes contain factual redundancy, for example from models that have been duplicated with minor modifications; the retriever frequently finds valid alternative evidence to synthesize the correct answer even if it misses the specific gold artifact ($E^*$). This reinforces that $R$ acts as a strict lower bound and suggests improved retrieval as a logical next step for metadata QA.
In summary, the wide gap between modest recall and reasoning accuracy suggests that navigating weakly linked metadata via dense retrieval is insufficient for reliable entity disambiguation and relation chaining.

\subsection{Retail Bank Data Lake: A failure of grounding, not retrieval}
\vspace{-4pt}
Table~\ref{tab:rag_results_financial} reports performance on the Retail Bank lake.
This domain is distinctive because benchmark items are framed as operational decisions requiring both unstructured policy retrieval and structured table lookups.
The \xmode{} baseline generally achieves high \emph{Policy Recall} (0.85+ for seven of the nine personas, though struggling on the Data Analyst and Salesperson workflows), but \emph{Table Recall} varies significantly across the board.
More importantly, final answer accuracy is often much lower than policy recall.
This exposes a synthesis gap: while agentic systems reliably retrieve abstract governing rules, they fundamentally struggle to ground those rules against specific operational ledgers to execute a faithful decision.
Notably, the zero-shot baseline achieves high accuracy on certain workflows (e.g., 0.88 for the Financial planner). While our semantic filter successfully removes simple factoid lookups, frontier LLMs possess strong parametric priors for standard financial principles, allowing them to often deduce correct operational outcomes without context. However, because this zero-shot reasoning lacks explicit provenance (Faithfulness is strictly zero), it highlights why our benchmark requires grounded synthesis; in an enterprise setting, an accurate but ungrounded answer remains an unauditable liability.
\begin{table}[t]
\centering
\small
\caption{\small Baseline results for the Retail Bank data lake. Metrics are Table Recall ($R_T$), Policy Recall ($R_P$), Accuracy ($A$), Faithfulness ($F$), and Token Count ($T$; k-tokens/query, as in Table~\ref{tab:rag_results_aiml}). Persona workflows span all four families in the task taxonomy depending on whether the task is isolated lookup, policy-ledger composition, multi-step trace, or anchor-conditioned policy selection.}
\label{tab:rag_results_financial}
\begin{tabularx}{\linewidth}{@{} l *{12}{>{\centering\arraybackslash}X} @{}}
\toprule
& \multicolumn{5}{c}{\textbf{\xmode{}}} & \multicolumn{5}{c}{\textbf{\modelresp{}}} & \multicolumn{2}{c}{\textbf{Zero-shot}} \\
\cmidrule(lr){2-6} \cmidrule(lr){7-11} \cmidrule(l){12-13}
\textbf{Persona} & $R_T$ & $R_P$ & $A$ & $F$ & $T$ & $R_T$ & $R_P$ & $A$ & $F$ & $T$ & $A$ & $T$ \\
\midrule
Accountant            & 0.79 & 0.94 & 0.60 & 0.45 & 1.15 & 0.95 & 1.00 & 0.73 & 0.48 & 1.20 & 0.25 & 0.19 \\
Auditor               & 0.84 & 0.90 & 0.58 & 0.51 & 1.00 & 0.94 & 0.85 & 0.58 & 0.55 & 1.12 & 0.69 & 0.19 \\
CS manager            & 0.70 & 0.87 & 0.70 & 0.67 & 1.05 & 0.86 & 1.00 & 0.88 & 0.80 & 1.03 & 0.69 & 0.13 \\
Customer service (CS) & 0.82 & 0.98 & 0.64 & 0.64 & 1.26 & 0.93 & 1.00 & 0.58 & 0.50 & 1.07 & 0.69 & 0.16 \\
Data analyst          & 0.62 & 0.69 & 0.59 & 0.49 & 1.10 & 0.89 & 0.65 & 0.55 & 0.38 & 1.04 & 0.31 & 0.16 \\
Financial planner     & 0.64 & 0.99 & 0.65 & 0.59 & 1.24 & 0.85 & 0.90 & 0.68 & 0.70 & 1.05 & 0.88 & 0.16 \\
Fraud analyst         & 0.72 & 0.92 & 0.61 & 0.47 & 1.10 & 0.95 & 0.90 & 0.68 & 0.60 & 1.27 & 0.81 & 0.15 \\
Insurance analyst     & 0.52 & 0.89 & 0.58 & 0.45 & 1.14 & 0.74 & 0.75 & 0.62 & 0.48 & 1.24 & 0.38 & 0.20 \\
Salesperson           & 0.56 & 0.66 & 0.44 & 0.42 & 1.15 & 0.77 & 0.60 & 0.30 & 0.23 & 1.09 & 0.38 & 0.15 \\
\bottomrule
\end{tabularx}
\vspace{-8pt}
\end{table}

\subsection{Drug Data Lake: Joint tabular QA remains challenging}
\vspace{-4pt}
Table~\ref{tab:rag_results_drug} reports performance on the Drug data lake, which stresses multimodal question answering across passages and tables.
The strongest results appear on Passage questions, where \xmode{} achieves near-perfect recall (0.98), accuracy (0.97), and faithfulness (0.97).
Passage-hop questions remain relatively strong: despite Passage Recall dropping to 0.72, answer accuracy is 0.86, suggesting systems can often synthesize correct answers from partially recovered text.
This gap indicates that retrieving the correct table is often not sufficient; the remaining challenge lies in extracting the right fields and composing them into a correct answer. Difficulty increases further for Table-hop questions, where Table Recall drops to 0.48 and accuracy to 0.44, showing that multi-step reasoning over structured biomedical evidence remains substantially harder than single-table lookup.

The mixed Passage-Table family further reinforces this point.
Although exact retrieval is very strong in both modalities---0.73 Passage Recall and 0.98 Table Recall---end-to-end accuracy reaches only 0.66.
In other words, the main bottleneck is no longer finding the relevant evidence, but correctly integrating heterogeneous evidence across text and tables.
Taken together, these results suggest that the Drug data lake is best characterized not as a retrieval-limited setting, but as a cross-modal synthesis setting in which joint reasoning over structured and unstructured biomedical evidence remains the main challenge.

\begin{table}[t]
\centering
\small
\caption{Baseline results for the Drug data lake. Metrics are Passage Recall ($R_P$), Table Recall ($R_T$), Accuracy ($A$), Faithfulness ($F$), and Token Count ($T$; k-tokens/query, as in Table~\ref{tab:rag_results_aiml}). Passage/Table map to Single-source lookups and anchor questions, Passage-Table maps to Cross-source, and Passage-hop/Table-hop map to Multi-hop. \modelresp{} token counts are estimated full generation tokens per question (approximate).}
\label{tab:rag_results_drug}
\begin{tabularx}{\linewidth}{@{} l *{12}{>{\centering\arraybackslash}X} @{}}
\toprule
& \multicolumn{5}{c}{\textbf{\xmode{}}} & \multicolumn{5}{c}{\textbf{\modelresp{}}} & \multicolumn{2}{c}{\textbf{Zero-shot}} \\
\cmidrule(lr){2-6} \cmidrule(lr){7-11} \cmidrule(l){12-13}
\textbf{Question Type} & $R_P$ & $R_T$ & $A$ & $F$ & $T$ & $R_P$ & $R_T$ & $A$ & $F$ & $T$ & $A$ & $T$ \\
\midrule
Passage       & 0.98 & --   & 0.97 & 0.97 & 0.66 & 1.00 & --   & 0.99 & 1.00 & 0.95 & 0.45 & 0.29 \\
Table         & --   & 0.81 & 0.60 & 0.76 & 0.59 & --   & 0.75 & 0.65 & 0.70 & 0.73 & 0.25 & 0.23 \\
Passage-Table & 0.73 & 0.98 & 0.66 & 0.82 & 0.66 & 0.90 & 1.00 & 0.79 & 0.85 & 0.86 & 0.47 & 0.26 \\
Passage-hop   & 0.72 & --   & 0.86 & 0.90 & 0.84 & 0.65 & --   & 0.86 & 0.90 & 1.30 & 0.57 & 0.28 \\
Table-hop     & --   & 0.48 & 0.44 & 0.62 & 0.67 & --   & 0.60 & 0.51 & 0.63 & 0.87 & 0.32 & 0.26 \\

\bottomrule
\end{tabularx}
\vspace{-8pt}
\end{table}

\subsection{Cross-Domain Takeaways}
\vspace{-4pt}
Across all three lakes, a consistent narrative emerges: \emph{high retrieval quality alone does not guarantee correct reasoning.} We highlight two major takeaways for future system design:

\textbf{The Divergence of Discovery vs. Synthesis:} By isolating source discovery from synthesis using a controlled context-bundle ablation (App~\ref{app:context_bundle_ablation}), we find that failure modes are strictly domain-dependent. When provided with perfect gold evidence, AI/ML accuracy saturates at 98.7\%, proving that the AI/ML lake is purely a \emph{discovery} bottleneck. Conversely, Drug lake accuracy only reaches 62.5\% even with perfect evidence, demonstrating that cross-modal reasoning over dense biomedical tables remains fundamentally broken in LLMs, independent of retrieval performance.

\textbf{The Compounding Cost and Latency of Agentic Navigation:} While standard single-shot RAG cannot natively navigate the relational topologies of the Retail Bank and Drug lakes, agentic loop-based methods (e.g., \xmode{}) introduce a compounding token and temporal overhead. Table~\ref{tab:rag_results_aiml} shows that the zero-shot baseline uses only \textasciitilde 0.05--0.07 k-tokens per query (\textasciitilde 0.06 on average). In contrast, single-shot RAG requires retrieving and processing substantially larger context, increasing token usage to \textasciitilde 0.34--0.49 k-tokens per query. Transitioning to iterative agentic navigation compounds this further. As reflected by the proxy token values in Tables~\ref{tab:rag_results_drug} and~\ref{tab:rag_results_financial}, iterative navigation increases token consumption significantly---in the Retail Bank lake, to roughly 1.00--1.27 k-tokens per query (about 5--8$\times$ over zero-shot and up to \textasciitilde 3$\times$ over standard RAG). Beyond token footprint, multi-step retrieval loops incur sequential latency bottlenecks, presenting a practical constraint for real-time operational environments like front-line customer service.

\section{Limitations}
\label{sec:limitations}

\textbf{Domain Scope and Synthetic Data:} \ours{} evaluates text and tables, excluding dynamic logs and raw audio/visual data (e.g., vision-model caveats rely on text metadata; see App~\ref{app:cases}). Furthermore, the Retail Bank domain relies on synthetic data for privacy, which may alter real-world noise distributions.

\textbf{Model Monoculture and Evaluator Bias:} Using an LLM to synthesize candidate questions risks introducing distributional bias that favors the generator's reasoning patterns. We mitigated this via strict human validation. However, evaluating baselines and rubric-based correctness with the same LLM family (GPT-5) used for synthesis introduces a risk of self-preference bias \citep{zheng2023judging}, which may overstate measured baseline performance. Future evaluations using independent frontier models will help separate general architectural bottlenecks from model-specific priors.

\textbf{Factual Redundancy:} Data lakes inherently contain redundant facts. Because \ours{} strictly tracks single-path gold evidence ($E^*_i$), systems might retrieve valid alternative provenance. Consequently, our retrieval metric serves as a conservative lower bound.

\section{Conclusion}
\label{sec:conclusion}
\vspace{-4pt}
We introduced \ours{}, a comprehensive benchmark and reproducible construction framework for grounded question answering over heterogeneous data lakes.
Encompassing AI/ML, retail banking, and biomedical domains, the benchmark provides a shared, human-validated evaluation environment that explicitly pairs natural language questions with precise, modality-aware evidence pointers.
Through baseline evaluations, we demonstrated that high retrieval quality alone is insufficient; modern retrieve-and-synthesize architectures fail differently across domains, struggling with relation chaining, policy grounding, and multimodal composition.
Ultimately, \ours{} provides the necessary evaluation infrastructure to drive the next generation of reliable, agentic QA systems capable of rigorous discovery and faithful attribution over real-world data lakes.

\section{Ethics Statement}

\ours{} includes retail banking and biomedical QA tasks where hallucinated or weakly grounded answers pose significant real-world risks.
Consequently, this benchmark explicitly evaluates \emph{Faithfulness} alongside accuracy, and human validation ensures that all queries are natively answerable from the retrieved context.
However, strong performance on \ours{} does not constitute an endorsement for the autonomous deployment of these systems in high-stakes, real-world decision pipelines.
\ours{} is intended as a rigorous evaluation resource, not a substitute for domain expertise or human oversight.

For the human validation, we recruited and paid data annotators using the Prolific platform \citep{prolific}, ensuring that research ethics were upheld. We acknowledge that ethical considerations are extremely important for any research involving human subjects, and we have ensured that our research meets the Code of Ethics for COLM 2026. Our research plan was reviewed and approved by the ethics review board at our institution. In addition, we took steps to ensure research ethics including: reviewing our institution's protocols for ethical research; only showing suitable content to participants; collecting consent from annotators regarding how the data would be collected, used, and retained; and paying all recruits a fair wage above the minimum wage in the jurisdiction of the recruitment platform.

\section{Reproducibility Statement}
To ensure the full reproducibility of our five-stage dataset construction pipeline and evaluation framework, we release all artifacts associated with this work. The \ours{} benchmark, including human validation annotations, explicit evidence pointers, and the underlying data lake corpora (subject to their respective licensing), is hosted on Github at \url{www.github.com/michael0402/LakeQuest}. Furthermore, our complete codebase---containing the automated generation pipeline, human-annotation UI, and all baseline implementations (RAG, \xmode{}, \modelresp{}) with exact prompts---is detailed in App~\ref{app:prompts} and will be open-sourced upon publication.

\section*{Acknowledgments}

We are grateful to the Vector Institute for their support of this work. This research was also supported in part by the Natural Sciences and Engineering Research Council of Canada (NSERC) through Discovery Grant.

\bibliographystyle{colm2026_conference}
\bibliography{references}

\clearpage

\appendix

\section{Detailed Dataset Statistics}
\label{app:statistics}

This section provides a granular breakdown of the human validation results across all three data lakes. As described in Sec~\ref{sec:human-eval}, all candidate question-answer pairs generated by the synthesis pipeline underwent rigorous human review. The tables below report the total number of synthesized candidates (\textbf{Questions}) alongside the number of instances that successfully passed three key human-annotation criteria:
\begin{itemize}[leftmargin=*, nosep]
    \item \textbf{Correct:} The reference answer is factually accurate and logically sound.
    \item \textbf{AIS (Attributable to Identified Sources):} The reference answer is strictly entailed by the provided evidence pointers, with no hallucinations or reliance on external knowledge.
    \item \textbf{Answerable:} The question provides sufficient context to be answered exclusively using the retrieved data lake artifacts.
\end{itemize}
Only instances that passed all criteria were included in the final benchmark.

\subsection{AI/ML Data Lake Statistics}
Table~\ref{tab:aiml_human_results} presents the validation breakdown for the AI/ML repository. In this domain, questions are structurally categorized based on the underlying metadata entities (Model Cards vs. Dataset Cards) and whether the reasoning requires multi-hop entity resolution across the AI/ML graph.

\begin{table}[htbp]
\centering
\small
\caption{\small Detailed human-annotation results for the AI/ML data lake, categorized by metadata entity types and graph topology.}
\label{tab:aiml_human_results}
\begin{tabularx}{\linewidth}{@{} >{\raggedright\arraybackslash}X >{\raggedleft\arraybackslash}X >{\raggedleft\arraybackslash}X >{\raggedleft\arraybackslash}X >{\raggedleft\arraybackslash}X @{}}
\toprule
\textbf{Question Type} & \textbf{Questions} & \textbf{Correct} & \textbf{AIS} & \textbf{Answerable} \\
\midrule
Model & 897 & 876 & 834 & 815 \\
Dataset & 975 & 925 & 929 & 896 \\
Model-Dataset & 623 & 592 & 570 & 562 \\
Model-hop & 741 & 701 & 706 & 683 \\
Dataset-hop & 699 & 656 & 648 & 622 \\
\midrule
\textbf{Total} & \textbf{3935} & \textbf{3750} & \textbf{3687} & \textbf{3578} \\
\bottomrule
\end{tabularx}
\end{table}

\subsection{Retail Bank Data Lake Statistics}
Table~\ref{tab:financial_human_results} outlines the validation results for the synthetic Retail Bank data lake. Unlike the other domains, Retail Bank QA instances are categorized by operational personas, testing how well models can ground their reasoning in strict policy constraints under different institutional workflows.

\begin{table}[htbp]
\centering
\small
\caption{\small Detailed human-annotation results for the Retail Bank data lake, categorized by persona-driven operational workflows.}
\label{tab:financial_human_results}
\begin{tabularx}{\linewidth}{@{} >{\raggedright\arraybackslash}p{3.5cm} >{\raggedleft\arraybackslash}X >{\raggedleft\arraybackslash}X >{\raggedleft\arraybackslash}X >{\raggedleft\arraybackslash}X @{}}
\toprule
\textbf{Persona} & \textbf{Questions} & \textbf{Correct} & \textbf{AIS} & \textbf{Answerable} \\
\midrule
Accountant & 549 & 478 & 461 & 454 \\
Auditor & 370 & 321 & 309 & 305 \\
Customer service (CS) & 328 & 306 & 311 & 303 \\
CS manager & 303 & 293 & 294 & 293 \\
Data analyst & 375 & 362 & 361 & 360 \\
Financial Planner & 228 & 216 & 215 & 215 \\
Fraud analyst & 416 & 391 & 381 & 378 \\
Insurance analyst & 429 & 378 & 376 & 366 \\
Salesperson & 314 & 302 & 295 & 290 \\
\midrule
\textbf{Total} & \textbf{3312} & \textbf{3047} & \textbf{3003} & \textbf{2964} \\
\bottomrule
\end{tabularx}
\end{table}

\subsection{Drug Data Lake Statistics}
Table~\ref{tab:drug_human_results} details the validation outcomes for the Drug data lake. The task families here stress cross-modal synthesis, dividing questions based on whether the required evidence resides in unstructured scientific text (Passages), structured biochemical records (Tables), or a combination of both.

\begin{table}[htbp]
\centering
\small
\caption{\small Detailed human-annotation results for the Drug data lake, categorized by modality-specific reasoning requirements.}
\label{tab:drug_human_results}
\begin{tabularx}{\linewidth}{@{} >{\raggedright\arraybackslash}X >{\raggedleft\arraybackslash}X >{\raggedleft\arraybackslash}X >{\raggedleft\arraybackslash}X >{\raggedleft\arraybackslash}X @{}}
\toprule
\textbf{Question Type} & \textbf{Questions} & \textbf{Correct} & \textbf{AIS} & \textbf{Answerable} \\
\midrule
Passage & 797 & 768 & 762 & 754 \\
Table & 776 & 739 & 720 & 701 \\
Passage-Table & 447 & 429 & 421 & 410 \\
Passage-hop & 706 & 681 & 668 & 646 \\
Table-hop & 848 & 817 & 802 & 793 \\
\midrule
\textbf{Total} & \textbf{3,574} & \textbf{3434} & \textbf{3373} & \textbf{3304} \\
\bottomrule
\end{tabularx}
\end{table}

\clearpage

\section{Prompt Templates}
\label{app:prompts}
This section details the exact prompt templates used for dataset synthesis, LLM-as-a-judge filtering, and baseline evaluation. 
\subsection{Candidate Synthesis Prompts}
\label{app:prompts_synthesis}
To generate candidate QA pairs, we provided the language model with a sampled evidence subgraph and the following instructions. The system prompt was shared across synthesis settings, while the task prompt was instantiated with minor wording changes depending on whether the evidence subgraph contained a model--dataset pair, multiple related model cards, multiple related dataset cards, or an anchor-plus-positive-plus-distractor setup.
\begin{mdframed}[backgroundcolor=gray!10, roundcorner=5pt, innertopmargin=10pt, innerbottommargin=10pt, skipabove=8pt, skipbelow=8pt, nobreak=true]
\small
\textbf{System Instructions:} \\
You are a data scientist working with models and data on huggingface.
\vspace{0.5cm}

\textbf{Input Context:} \\
\{evidence\_subgraph\}
\vspace{0.5cm}

\textbf{Task:} \\
I'm going to show you one or more model cards and/or dataset cards. What natural question and answer might you ask that implicitly requires consulting information from the shown evidence subgraph? You should pretend that you cannot see the cards when asking the question (except in the anchored setting, where the anchor card is visible but the supporting and distractor cards are not). If you cannot think of one, then return an empty string. This should be a question that naturally requires consulting information from the provided evidence and should not be a chain of multiple unrelated questions that merely happen to require multiple cards. After the question, provide: (1) a direct answer (short, exact answer), and (2) an explained answer (slightly longer answer with brief justification from the cards). Always answer in the following format:
\begin{verbatim}
<question>
<direct_answer>
<explained_answer>
\end{verbatim}
\end{mdframed}

\subsection{Baseline Evaluation Prompts: Agentic Tool Use (\xmode{})}
\label{app:prompts_agent}
For our agentic baselines, we framed the data lake navigation as an iterative tool-use task.

\begin{mdframed}[backgroundcolor=blue!5, roundcorner=5pt, innertopmargin=10pt, innerbottommargin=10pt, skipabove=8pt, skipbelow=8pt, nobreak=true]
\small
\textbf{Agent Instructions:} \\
You are an AI agent for data lake question answering. Given a user question, iteratively reason over the question, retrieve the top-$k$ relevant documents from the indexed data lake for the current question or sub-question, and summarize each document with respect to both the original question and the current sub-question. After each retrieval round, decide whether the accumulated evidence is sufficient to answer the original question; if not, propose the next sub-question and continue. We use up to 3 iterations, retrieve $k=5$ documents per step, and use either similarity search or MMR retrieval. After the final step, generate the answer only from the accumulated evidence.
\end{mdframed}

\clearpage

\section{Human Annotation Guidelines and Interface}
\label{app:annotation}
To ensure high data quality, annotators were provided with a custom web interface that displayed the generated question, the synthesized answer, and the exact supporting evidence chunks. 
Annotators were tasked with independently verifying two criteria: \emph{Answerability} (can the question be answered exclusively using the cited evidence?) and \emph{Attribution} (is the generated answer completely faithful to the cited text/table?).
We emphasize that this task evaluates strict textual entailment and reading comprehension based entirely on the provided evidence, rather than requiring annotators to verify external scientific or financial validity.
\begin{figure}[htbp]
\centering
\includegraphics[width=\linewidth]{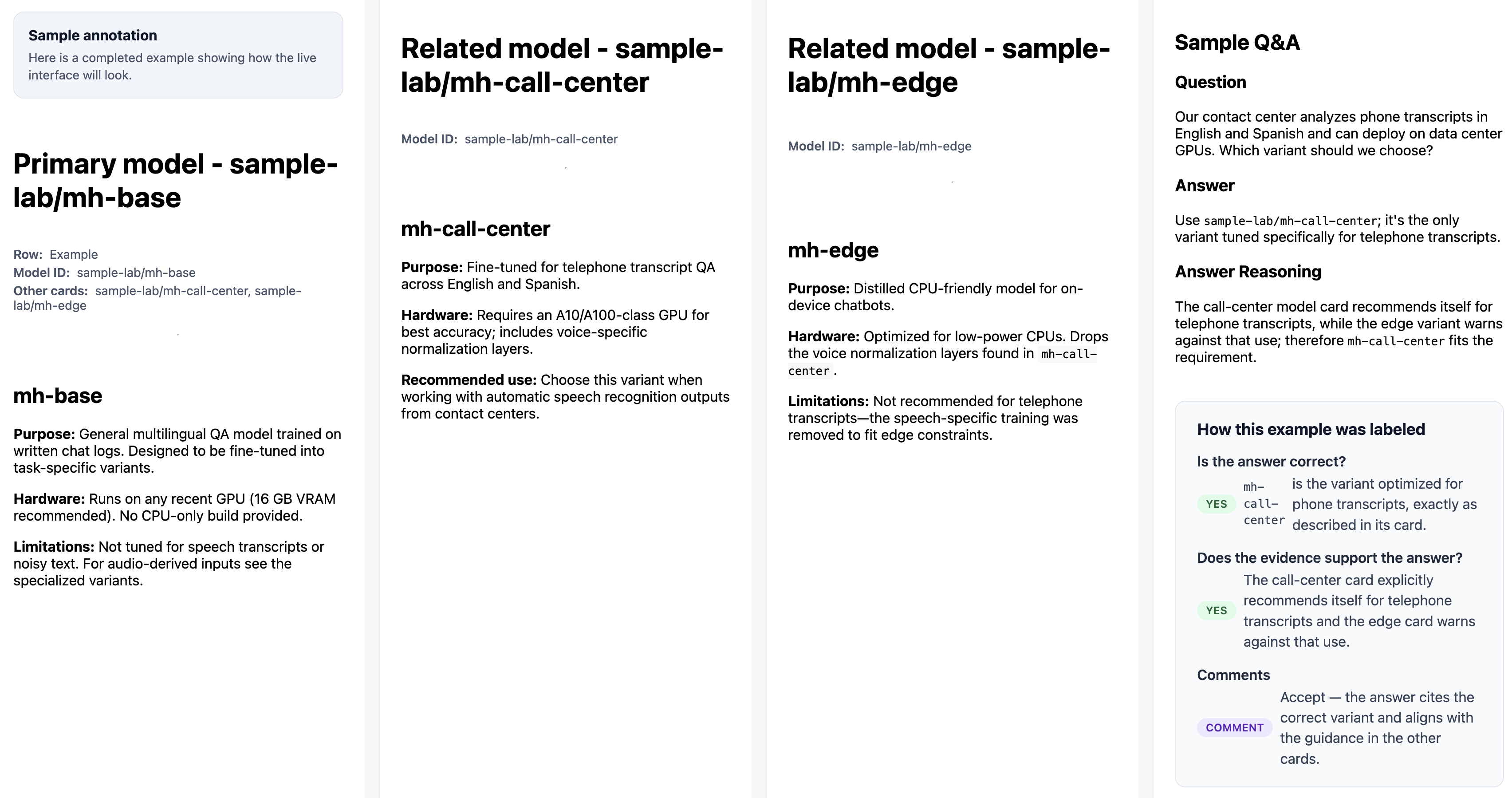}
\caption{\small Screenshot of the custom validation UI. Annotators are presented with the highlighted evidence constraints and must explicitly accept or reject the instance based on the provided rubrics.}
\label{fig:annotation_ui}
\end{figure}
\begin{itemize}
\item \textbf{Is the answer correct?} Mark ``Yes'' only when the answer matches facts stated in the provided evidence.
\item \textbf{Does the evidence support the answer?} Mark ``Yes'' only when specific sentences/rows in the provided evidence explicitly support the answer.
\item \textbf{Any additional comments/thoughts.} Add a short note explaining the decision, especially for ``No'' labels, and reference the relevant evidence span.
\item \textbf{Use only provided evidence.} Do not use outside knowledge; if the answer contradicts the provided evidence, mark it incorrect and unsupported.
\end{itemize}

\clearpage

\section{Extended Case Studies and Error Analysis}
\label{app:cases}

To qualitatively illustrate the quantitative failure modes discussed in Sec~\ref{sec:results}, we provide representative trace-level examples from the AI/ML and Drug data lakes. The cases below show three recurring error patterns: schema-confusable retrieval, relation chaining failure across closely related model cards, and multimodal metadata composition failure in vision--language datasets.

\begin{table}[htbp]
\centering
\small
\caption{\small A representative schema-grounding failure in the AI/ML data lake: retrieval finds the right capability class (function calling) but misses the target card's exact token schema.}
\label{tab:error_case_ai_schema}
\begin{tabularx}{\linewidth}{@{} >{\hsize=0.5\hsize}X >{\hsize=1.5\hsize}X @{}}
\toprule
\textbf{Domain} & \textbf{AI/ML (Schema-Confusable Retrieval)} \\
\midrule
\textbf{Question} & What prompt format does \texttt{0-hero/Matter-0.1-7B} use, and which special tokens does it provide for function calling? \\
\midrule
\textbf{Gold Evidence} & The \texttt{Matter-0.1-7B} card states that the model uses ChatML prompt format; its Function Calling section lists \texttt{<|begin\_func|>}, \texttt{<|end\_func|>}, \texttt{<|begin\_func\_response|>}, and \texttt{<|end\_func\_response|>}. \\
\midrule
\textbf{Gold Answer} & ChatML prompt format; function-call tokens \texttt{<|begin\_func|>} and \texttt{<|end\_func|>}; function-response tokens \texttt{<|begin\_func\_response|>} and \texttt{<|end\_func\_response|>}. \\
\midrule
\textbf{RAG Predicted Evidence} & The retriever surfaced schema-similar tool-use models such as \texttt{meetkai/functionary-medium-v3.1} and \texttt{meetkai/functionary-small-v3.1}, plus unrelated cards, instead of grounding on the target Matter card. \\
\midrule
\textbf{RAG Predicted Answer} & ``Prompt format: ChatML-style. Function-calling special tokens: \texttt{<|tool\_call|>} and \texttt{<|tool\_response|>}.'' \\
\midrule
\textbf{Failure Mode Analysis} & The retriever found semantically similar function-calling cards, but not the target-specific token schema. The generator then copied the more common \texttt{tool\_call}/\texttt{tool\_response} convention, yielding a partially correct yet still wrong answer. \\
\bottomrule
\end{tabularx}
\end{table}

\begin{table}[htbp]
\centering
\small
\caption{\small A relation-chaining failure: the system retrieves locally similar model cards but does not gather the exact sibling set required for accurate cross-card comparison.}
\label{tab:error_case_ai_relation}
\begin{tabularx}{\linewidth}{@{} >{\hsize=0.5\hsize}X >{\hsize=1.5\hsize}X @{}}
\toprule
\textbf{Domain} & \textbf{AI/ML (Relation Chaining)} \\
\midrule
\textbf{Question} & Among fsicoli's fine-tuned Whisper models (\texttt{whisper-small-pt-1000h}, \texttt{whisper-medium-pt-1000h}, and \texttt{whisper-large-v3-pt-1000h}), which achieves the lowest WER on the shared evaluation dataset, and what are the WERs for each model? \\
\midrule
\textbf{Gold Evidence} & The three model cards report WERs of 0.1490 (\texttt{whisper-small-pt-1000h}), 0.1147 (\texttt{whisper-medium-pt-1000h}), and 0.1113 (\texttt{whisper-large-v3-pt-1000h}), all on the same Portuguese speech dataset family. \\
\midrule
\textbf{Gold Answer} & \texttt{whisper-large-v3-pt-1000h} has the lowest WER (0.1113), followed by \texttt{whisper-medium-pt-1000h} (0.1147) and \texttt{whisper-small-pt-1000h} (0.1490). \\
\midrule
\textbf{RAG Predicted Evidence} & The retriever pulled neighboring Whisper cards such as \texttt{fsicoli/whisper-large-v3-pt-cv16} and several unrelated Whisper models, but failed to assemble the three required \texttt{pt-1000h} cards into a single comparison set. \\
\midrule
\textbf{RAG Predicted Answer} & ``I don't see WERs for fsicoli's \texttt{pt-1000h} models in the provided context. The only fsicoli metric here is for \texttt{whisper-large-v3-pt-cv16} with WER \(\approx\) 0.1076.'' \\
\midrule
\textbf{Failure Mode Analysis} & This error is not a pure hallucination; it is a chaining failure. The system retrieved a nearby family member with a very similar identifier, but missed the exact trio needed for cross-card aggregation, so it could not complete the comparison. \\
\bottomrule
\end{tabularx}
\end{table}

\begin{table}[htbp]
\centering
\small
\caption{\small A multimodal composition failure: the answer depends on jointly grounding license metadata and image-rights caveats across two vision--language dataset cards.}
\label{tab:error_case_ai_multimodal}
\begin{tabularx}{\linewidth}{@{} >{\hsize=0.5\hsize}X >{\hsize=1.5\hsize}X @{}}
\toprule
\textbf{Domain} & \textbf{AI/ML (Multimodal Composition)} \\
\midrule
\textbf{Question} & Do \texttt{FreedomIntelligence/ALLaVA-4V} and \texttt{FreedomIntelligence/ALLaVA-4V-Chinese} allow commercial use? \\
\midrule
\textbf{Gold Evidence} & The \texttt{ALLaVA-4V} dataset card lists license \texttt{cc-by-nc-4.0}; \texttt{ALLaVA-4V-Chinese} lists \texttt{apache-2.0}. The \texttt{ALLaVA-4V} card also states that the authors ``do not own right to any image contained within the \texttt{images.zip} file,'' requiring users to respect original image copyrights. \\
\midrule
\textbf{Gold Answer} & No. \texttt{ALLaVA-4V} is non-commercial, while \texttt{ALLaVA-4V-Chinese} is Apache-2.0; moreover, commercial reuse still requires checking the rights of the underlying third-party images. \\
\midrule
\textbf{RAG Predicted Evidence} & The retriever returned image-generation model cards such as \texttt{LarryAIDraw/*} and \texttt{svjack/Stable-Diffusion-FineTuned-zh-*}, rather than the two ALLaVA dataset cards whose licenses had to be composed. \\
\midrule
\textbf{RAG Predicted Answer} & ``I can't confirm that from the provided context. If the license is Apache-2.0, MIT, BSD, or CC-BY, commercial use is generally allowed; if it is CC-BY-NC, it is not allowed.'' \\
\midrule
\textbf{Failure Mode Analysis} & The query requires composing metadata from two vision--language dataset cards and incorporating an image-rights caveat. Retrieval drifted toward visually related generative model cards, so the generator defaulted to generic licensing heuristics instead of grounding on the relevant multimodal datasets. \\
\bottomrule
\end{tabularx}
\end{table}

\begin{table}[htbp]
\centering
\small
\caption{\small Representative Drug failure case from the evidence-provided ablation: the model receives the correct table evidence but still fails to produce the correct answer.}
\label{tab:error_case_drugbank_generation}
\begin{tabularx}{\linewidth}{@{} >{\hsize=0.5\hsize}X >{\hsize=1.5\hsize}X @{}}
\toprule
\textbf{Domain} & \textbf{Drug (Evidence-Provided Generation Failure)} \\
\midrule
\textbf{Question} & Which cited reference mentions ``Lorazepam'', and does that reference include an ISBN? \\
\midrule
\textbf{Gold Evidence} & In \texttt{drugbank-drugs\_textbooks}, the row with \texttt{ref-id=T385} has citation ``Ghiasi N. and Marwaha R. (2018). Lorazepam'' and \texttt{isbn = NaN} (\texttt{parent\_key=DB00186}). \\
\midrule
\textbf{Gold Answer} & \texttt{T385}; it does not include an ISBN (missing / NaN). \\
\midrule
\textbf{LLM Generated Answer} & ``Not determinable from the provided snippets; no citation mentioning `Lorazepam' is shown, so the ISBN status is unknown.'' \\
\midrule
\textbf{Failure Mode Analysis} & In the context-bundle ablation, the model was given the correct evidence chunk (plus four distractor tables), so retrieval was not the bottleneck. The key row was present in context, but the model failed to attend to and extract it, defaulting to an abstention-style response. This is a generation-under-sufficient-evidence failure rather than a retrieval miss. \\
\bottomrule
\end{tabularx}
\end{table}

\clearpage

\section{Controlled Context-Bundle Ablation}
\label{app:context_bundle_ablation}

We ran a fixed context-bundle ablation on the AI/ML, Retail Bank, and Drug data lakes to reduce retrieval variance and better isolate answer synthesis. In each case, the model was given a fixed evidence bundle containing the gold evidence plus relevant distractors, and we evaluated both a direct answer and a reasoning answer using the same LLM-as-a-judge pipeline as in the main experiments.

\begin{table}[htbp]
\centering
\small
\caption{\small LLM correctness results for the fixed context-bundle ablation across data lakes.}
\label{tab:context_bundle_ablation}
\setlength{\tabcolsep}{4pt}
\begin{tabularx}{\linewidth}{@{} l
l
>{\raggedleft\arraybackslash}X
>{\raggedleft\arraybackslash}X
>{\raggedleft\arraybackslash}p{3.5cm} @{}}
\toprule
\textbf{Data Lake} & \textbf{Baseline} & \textbf{Avg. Correctness} & \textbf{Direct Correctness} & \textbf{Reasoning Correctness} \\
\midrule
AI/ML & Context Bundle & 0.9875 & 0.9750 & 1.0000 \\
Retail Bank & Context Bundle & 0.7500 & 0.7500 & 0.7500 \\
Drug & Context Bundle & 0.6250 & 0.6000 & 0.6500 \\
\bottomrule
\end{tabularx}
\end{table}

The results show a sharp cross-domain difference. On AI/ML, performance is nearly saturated once gold evidence is present; Retail Bank is materially higher but not saturated, while Drug remains lowest under the same setup, indicating that synthesis over heterogeneous evidence remains a substantial challenge; reasoning is equal to or stronger than direct answers across lakes.

\clearpage

\section{LLM-as-a-Judge Consistency Analysis}
\label{app:llm_judge_consistency}

To assess the stability of our rubric-based LLM-as-a-judge evaluation, we repeated the same AI/ML data lake evaluation three times under an identical setup and report per-question-family correctness in Table~\ref{tab:aiml_llm_judge_consistency}. The results are highly consistent across runs, with only minor variation across the three repeated tests, suggesting that the judge-based correctness metric is sufficiently stable for the comparative analyses reported in the main paper.

\begin{table}[htbp]
\centering
\small
\caption{\small Repeated-run LLM correctness on the AI/ML data lake.}
\label{tab:aiml_llm_judge_consistency}
\setlength{\tabcolsep}{4pt}
\begin{tabularx}{\linewidth}{@{} l
>{\raggedleft\arraybackslash}X
>{\raggedleft\arraybackslash}X
>{\raggedleft\arraybackslash}X @{}}
\toprule
\textbf{Question Type} & \textbf{LLM Correctness -- Test 1} & \textbf{LLM Correctness -- Test 2} & \textbf{LLM Correctness -- Test 3} \\
\midrule
Model & 0.250 & 0.260 & 0.245 \\
Dataset & 0.140 & 0.145 & 0.140 \\
Model-Dataset & 0.320 & 0.340 & 0.320 \\
Model-hop & 0.155 & 0.165 & 0.160 \\
Dataset-hop & 0.380 & 0.385 & 0.375 \\
\bottomrule
\end{tabularx}
\end{table}

\clearpage

\section{Retail Bank Data Lake QA Generation Example}
\label{app:finlake_qa_generation}

\begin{figure}[htbp]
    \centering
    \includegraphics[width=\linewidth]{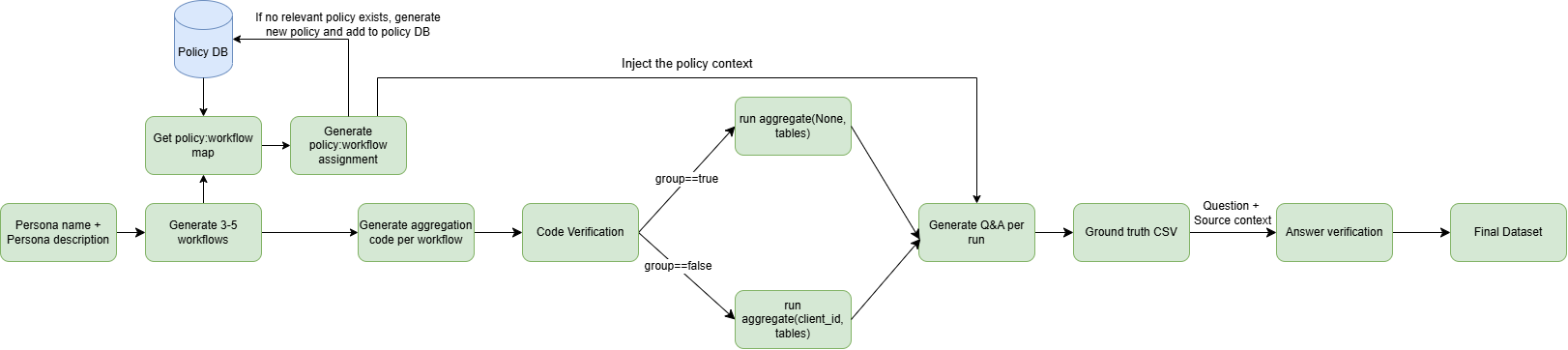}
    \caption{\small Example of QA generation for the Retail Bank data lake.}
    \label{fig:finlake_qa_generation}
\end{figure}

Figure~\ref{fig:finlake_qa_generation} illustrates the end-to-end question-answer synthesis pipeline for the Retail Bank data lake. The process begins by sampling a dense evidence subgraph consisting of a client profile, a transaction ledger, and a governing institutional policy. The generator model is instructed to synthesize a realistic operational query (e.g., a fraud analyst checking a transaction limit) that strictly requires joining facts across all three sampled artifacts. Following generation, the pipeline programmatically filters deictic or ill-formed queries, and an LLM-as-a-judge verifies that the question cannot be answered via parametric memory alone. Finally, the instance undergoes human validation to ensure strict attribution to the highlighted evidence spans.

\clearpage

\section{Open Source Model Results Ablation}

\begin{table}[htbp]
\centering
\small
\caption{\small Baseline results for the AI/ML data lake using gpt-oss-20b}
\label{tab:rag_results_aiml_gptoss20b}
\begin{tabularx}{\linewidth}{@{} l *{6}{>{\centering\arraybackslash}X} @{}}
\toprule
& \multicolumn{4}{c}{\textbf{RAG}} & \multicolumn{2}{c}{\textbf{Zero-shot}} \\
\cmidrule(lr){2-5} \cmidrule(l){6-7}
\textbf{Question Type} & $R$ & $A$ & $F$ & $T$ & $A$ & $T$ \\
\midrule
Model         & 0.09 & 0.29 & 0.21 & 0.36 & 0.20 & 0.05 \\
Dataset       & 0.15 & 0.09 & 0.11 & 0.31 & 0.07 & 0.05 \\
Model-Dataset & 0.12 & 0.30 & 0.25 & 0.35 & 0.13 & 0.05 \\
Model-hop     & 0.13 & 0.17 & 0.12 & 0.43 & 0.14 & 0.06 \\
Dataset-hop   & 0.23 & 0.28 & 0.30 & 0.44 & 0.18 & 0.06 \\
\bottomrule
\end{tabularx}
\vspace{-8pt}
\end{table}

\begin{table}[htbp]
\centering
\small
\caption{\small Baseline results for the Retail Bank data lake using gpt-oss-20b. }
\label{tab:rag_results_financial_gptoss20b}
\begin{tabularx}{\linewidth}{@{} l *{12}{>{\centering\arraybackslash}X} @{}}
\toprule
& \multicolumn{5}{c}{\textbf{\xmode{}}} & \multicolumn{5}{c}{\textbf{\modelresp{}}} & \multicolumn{2}{c}{\textbf{Zero-shot}} \\
\cmidrule(lr){2-6} \cmidrule(lr){7-11} \cmidrule(l){12-13}
\textbf{Persona} & $R_T$ & $R_P$ & $A$ & $F$ & $T$ & $R_T$ & $R_P$ & $A$ & $F$ & $T$ & $A$ & $T$ \\
\midrule
Accountant            & 0.79 & 0.94 & 0.50 & 0.37 & 1.02 & 0.95 & 1.00 & 0.61 & 0.40 & 1.08 & 0.18 & 0.15 \\
Auditor               & 0.84 & 0.90 & 0.49 & 0.42 & 0.92 & 0.94 & 0.85 & 0.50 & 0.47 & 1.00 & 0.56 & 0.16 \\
Customer service (CS) & 0.82 & 0.98 & 0.55 & 0.53 & 1.12 & 0.93 & 1.00 & 0.49 & 0.39 & 0.96 & 0.56 & 0.14 \\
CS manager            & 0.70 & 0.87 & 0.61 & 0.56 & 0.95 & 0.86 & 1.00 & 0.77 & 0.69 & 0.93 & 0.57 & 0.11 \\
Data analyst          & 0.62 & 0.69 & 0.50 & 0.40 & 1.00 & 0.89 & 0.65 & 0.46 & 0.31 & 0.95 & 0.24 & 0.13 \\
Financial planner     & 0.64 & 0.99 & 0.56 & 0.50 & 1.10 & 0.85 & 0.90 & 0.58 & 0.60 & 0.96 & 0.72 & 0.14 \\
Fraud analyst         & 0.72 & 0.92 & 0.52 & 0.39 & 0.99 & 0.95 & 0.90 & 0.59 & 0.50 & 1.14 & 0.67 & 0.13 \\
Insurance analyst     & 0.52 & 0.89 & 0.49 & 0.37 & 1.02 & 0.74 & 0.75 & 0.54 & 0.39 & 1.10 & 0.29 & 0.17 \\
Salesperson           & 0.56 & 0.66 & 0.35 & 0.33 & 1.01 & 0.77 & 0.60 & 0.23 & 0.18 & 0.97 & 0.29 & 0.13 \\
\bottomrule
\end{tabularx}
\vspace{-8pt}
\end{table}

\begin{table}[htbp]
\centering
\small
\caption{Baseline results for the Drug data lake using gpt-oss-20b.}
\label{tab:rag_results_drug_gptoss20b}
\begin{tabularx}{\linewidth}{@{} l *{12}{>{\centering\arraybackslash}X} @{}}
\toprule
& \multicolumn{5}{c}{\textbf{\xmode{}}} & \multicolumn{5}{c}{\textbf{\modelresp{}}} & \multicolumn{2}{c}{\textbf{Zero-shot}} \\
\cmidrule(lr){2-6} \cmidrule(lr){7-11} \cmidrule(l){12-13}
\textbf{Question Type} & $R_P$ & $R_T$ & $A$ & $F$ & $T$ & $R_P$ & $R_T$ & $A$ & $F$ & $T$ & $A$ & $T$ \\
\midrule
Passage       & 0.98 & --   & 0.90 & 0.91 & 0.58 & 1.00 & --   & 0.93 & 0.94 & 0.84 & 0.34 & 0.24 \\
Table         & --   & 0.81 & 0.50 & 0.65 & 0.52 & --   & 0.75 & 0.55 & 0.60 & 0.64 & 0.18 & 0.19 \\
Passage-Table & 0.73 & 0.98 & 0.56 & 0.70 & 0.58 & 0.90 & 1.00 & 0.68 & 0.75 & 0.75 & 0.36 & 0.22 \\
Passage-hop   & 0.72 & --   & 0.74 & 0.79 & 0.74 & 0.65 & --   & 0.75 & 0.80 & 1.12 & 0.43 & 0.24 \\
Table-hop     & --   & 0.48 & 0.35 & 0.50 & 0.58 & --   & 0.60 & 0.41 & 0.52 & 0.75 & 0.23 & 0.21 \\
\bottomrule
\end{tabularx}
\vspace{-8pt}
\end{table}

\end{document}